# Bayesian Grasp：Robotic visual stable grasp based on prior tactile knowledge

Teng Xue, Wenhai Liu, Mingshuo Han, Zhenyu Pan, Jin Ma, Quanquan Shao, Weiming Wang*

*Abstract*— Robotic grasp detection is a fundamental capability for intelligent manipulation in unstructured environments. Existing work mainly employs visual and tactile fusion to achieve stable grasp, while, the whole process depends heavily on regrasping, which calls for much time to regulate and evaluate. We propose a novel way to improve robotic grasping: by using learned tactile knowledge, a robot can achieve stable grasp through simple visual input.

First, we construct a prior tactile knowledge learning framework with a novel grasp quality metric determined by measuring its resistance to external perturbations. Second, we propose a multi-phase Bayesian Grasp architecture to generate stable grasp configurations through a single RGB image based on learned tactile knowledge.

Results show that this framework can classify the outcome of grasps with an average accuracy of 86% on known objects and 79% on novel objects. The prior tactile knowledge improves the successful rate of 55% over traditional vision-based strategies.

## I. INTRODUCTION

With the development of robotics and multi-modal technology, robots are able to acquire more and more human-like perception and abilities. However, they are still far from resembling the way humans manipulate objects, especially for stably grasping.

Vision plays an important role in the grasp planning stage and proper grasp configurations can be generated based on global visual features [1]. However, grasping is a physical contact process related to object attributes, hand configurations, and frictions. A seemingly stable grasp configuration predicted based on visual information may be influenced by physical contact during manipulation and fail in the execution stage.

To better reflect the physical contact between hands and target objects, tactile sensing is introduced to make up the deficiency of vision, which can evaluate the subtle changes of pressure values, even objects position.

Di Guo et al. [2] proposed a hybrid deep architecture fusing visual and tactile sensing to generate stable grasp configurations. Emil Hyttinen [3], MM. Zhang [4], Deen Cockburn [5] and Hogan F R et al. [6] employed tactile data into current grasp to evaluate hypothetical configurations and guide the robot to regrasp by combining the stability estimation method and the grasp adaption method.

However, these methods depend on regrasping many times, which no doubt needs much time to regulate and evaluate, again and again, until satisfying the stability requirement.

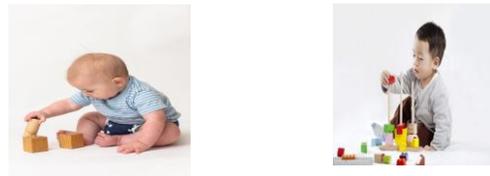

(a)　　　　　　　　(b)
Fig. 1.　Illustration of stable grasp at different stage of humans

Imagine the difference between an infant and a boy attempting to grasp stably. Figure 1 (a) depicts an example that an infant tries to grasp toy blocks. It's easy to find that he has no idea of where to grasp, and the common results are falling or slipping. And figure 1 (b) shows a grown-up boy can easily build great works with blocks. As Melvyn A. Goodale points out in [7], the reason is that infants cannot justify which point is appropriate to grasp at the first moment of looking at the scene for lack of massive tactile experience, which can be seen as a kind of prior tactile knowledge.

Therefore, we think that the regrasp policy delays robots staying at the infant stage, and we want to endow robots with prior tactile knowledge to act like a grown-up boy. To achieve this target, we propose a novel grasp mode, Bayesian Grasp (see Fig. 2), which applies much prior knowledge into the final execution process, and we focus on prior tactile knowledge mainly in this paper.

The contribution of this paper can be concluded as follows:

1) Grasp quality Metric. The metric shows great robustness for its independence of intensive spatio-temporal tactile data acquisition, and only a few contact points are required to achieve quality evaluation.

2) Bayesian Grasp Policy. A multi-phase architecture is proposed for stably grasping, which can generate ideal grasp configuration from a single RGB image based on learned tactile knowledge.

To our knowledge, this work is the first to present a framework to learn prior tactile knowledge and apply it to generate stable grasp configuration, which imitates humans' advanced intelligence during grasping.

*Resrach supported by NaturalScience Foundation of China (51775332, 51675329) Foundation.

Teng Xue, Wenhai Liu, Mingshuo Han, Jin Ma, Quanquan Shao and Weiming Wang are with the Institute of Knowledge Based Engineering, School of Mechanical Engineering, Shanghai Jiao Tong University, Shanghai, 200240, China(corresponding author phone: 008621-34206501; fax: 008621-34206840; e-mail: wangweiming@ sjtu.edu.cn, {xueteng, sjtu-wenhai, mrjay101, pan13855965084, heltonma, shao124}@sjtu.edu.cn).

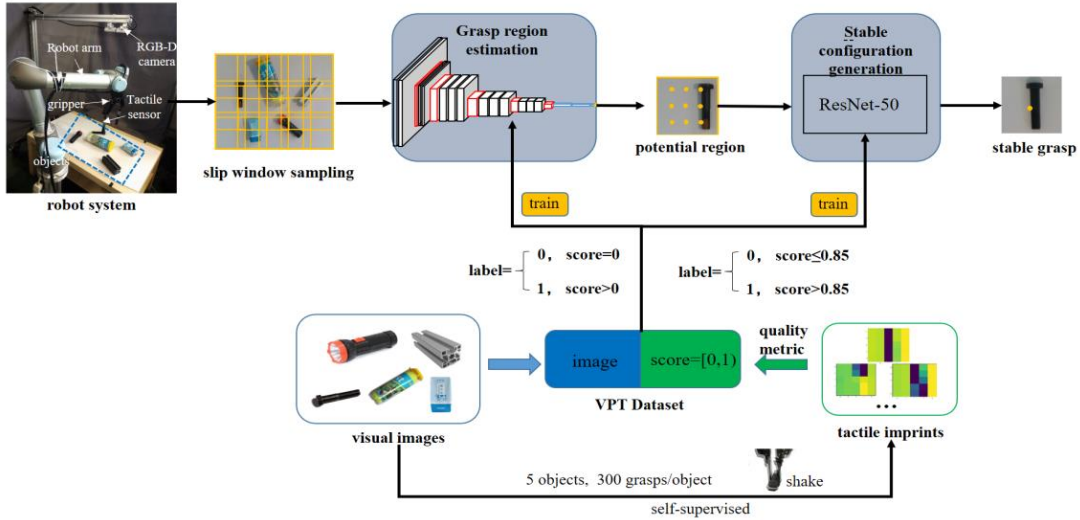

Fig. 2. **Overview** of our system and Bayesian Grasp policy. Our robot arm operates over a workspace observed by a statically mounted RGB-D camera. A visual and tactile joint dataset, VPT Dataset, is built with 5 objects and 3000 data after data augmentation. The visual image is re-projected into 100×100 patches through slipping window sampling. The little patches are then fed into Grasp Region Estimation (GRE) Network, which is trained on the VPT Dataset with redefined label, to determine potential grasp region. And then the potential patch is fed into Stable Configuration Generation (SCG) Network to generate stable grasp configuration with accurate point coordinates and angle. Both GRE and SCG network are trained based on the VPT Dataset, but with redefined labels.

## II. RELATED WORK

### A. Stable grasp policy

Generally, robotic grasp is just the first step of complex manipulation, and a stable grasp is essential to complete the next operations, such as fine assembly, hand over task and so on. A closely related work to our approach is [6], Hogan et al. proposed a data-driven grasp quality metric and a model –based regrasp policy, which introduced the learned grasp quality model to plan robust regrasps. This method avoids learning directly from generous raw data and shows high accuracy in novel objects recognition. However, the regrasp policy needed to adjust the contact point continuously following the predicted scores, rather than grasping stably at the first glance of the object. Similarly, Emil Hyttinen et al. [3] proposed a binary stability classifier to compute the probability of stable grasp in the vicinity of an already applied grasp. Grasp configurations were then changed constantly with the predicted estimated stable probability. Di Guo et al. [2] combined visual and tactile sensing for robotic grasp detection. After the initial grasp rectangle was generated under visual sensing, tactile information and strain gauge were also used to calculate the grasp assessment value. Grasp was repeated continuously until the grasp value exceeds the pre-set threshold. Roberto Calandra et al. [10] proposed an end-to-end action-conditional model that predicted the outcome of a candidate grasp adjustment, and executed a grasp by iteratively selecting the most promising action.

Generally speaking, existing researches commonly adopted regrasp policy as a stable grasping strategy. Although focusing on selecting grasp adjustments improves the grasp a lot contrary to estimating the stability of ongoing grasp [11][12], regrasp policy depends heavily on iteratively adjusting many times, which means that stable grasp configuration cannot be generated at the first sight of the scenario, and this action is far away from the way of stably grasping of human beings.

### B. Grasp quality metric

To assess grasp quality, varieties of metrics have been proposed considering the attributes of tactile sensors. Roberto Calandra et al. [10] used two GelSight sensors and an RGB camera to acquire raw visuo-tactile information. A binary grasp outcome, successful grasp or unsuccessful grasp, was defined as the result of an executing action at the current state. Furthermore, Peter K. Allen et al. [8] equipped the robot arm with a Barrett hand. They utilized forward kinematics to determine the location and orientation of each sensor cell and used a bag-of-words model to learn grasp stability, which was also a binary classifier, stable or unstable. Besides, Hogan et al. [6] assessed the grasp quality by shaking the end-effector after grasping and the tactile sensor was called as GelSlim, an improved version based on Gelsight. A metric, a continuous value from 0 to 1, was defined to evaluate the ability of the grasp towards external forces. And the score corresponded to 3 conditions: failure, falling and success.

Considering the unstable situation, Slip detection plays an important role in robotic dexterous manipulation, [13] has given a brief review of the transduction principles and technological approaches of slip detection. Howe and Cutkosky [14] pointed out that sensing accelerations should be used to detect displacement changes and a tactile sensor with an accelerator under the thin rubber skin is designed to detect the slip or incipient slip state of objects. Melchiorri et al. [15] measured the normal and tangential forces of the contact point together and the ratio of these two factors was compared to the frictional coefficient of the surface based on Coulomb friction model to predict slip. Su and Chebotar et al. [16][17] used BioTac, a biomimetic tactile sensor consisting of a rigid core housing an array of 19 electrodes, to detect slip through measuring the change in tangential force and vibration, and they also employed a Spatio-temporal feature descriptor to extract the features of BioTacs in different grasp situation and Support Vector Machine with a linear kernel was used for slip prediction classifier learning.

Fernandez et al. [18] designed a low-cost and durable force sensor to detect structural micro-vibrations of high frequencies in real-time and, therefore, slipping contacts. James et al. [19] measured the positions of internal pins embedded in a biomimetic tactile sensor and then used a support vector machine to discriminate static and slipping objects. Wenzhen Yuan et al. [20] developed GelSight tactile sensor by adding markers on the gel surface to sense the normal, shear and torsional load on the contact surface based on the elastomer's deformation during contact and Siyuan Dong et al. [21] improved the GelSight system using a Lambertian membrane and new illumination system, which could measure the relative displacement between object and markers. Jianhua Li et al. [22] used GelSight tactile sensor and an external camera to capture information during the lifting process. Sequenced data was fed into a trained DNN to classify a grasp to be stable or slippery.

Totally, the commonly used tactile sensors are array-based pressure sensors and vision-based optical tactile sensors, and the mostly used are BioTac (Syntouch llc, Los Angeles, CA, USA), Barrette Hand (Barrett Technologies, Cambridge, MA, USA) and GelSight [23] et al., which can provide more information but are too expensive or difficult to manufacture. It is meaningful to explore a novel grasp quality metric based on simple tactile sensors and slight perceptual information for potential large-scale application in complex extreme unstructured environments, where sensors cannot acquire enough perception and low cost is required.

III. PRIOR TACTILE KNOWLEDGE LEARNING

In this section, we build a novel tactile-based grasp quality metric that better adapts to the physical characteristics of our grippers. Then, we use this grasp quality metric to evaluate each image region with grasp score, which is a process of learning prior tactile knowledge.

*A. Grasp quality metric*

Robots can grasp an object but cannot evaluate the grasp quality through vision only, especially distinguishing the state as slippery or stable. In this paper, we propose a novel grasp quality metric. The quality is evaluated through a continuous score from 0 to 1. The higher the score is, the more stable the grasp is. And Fig. 2. shows the tactile images and quality scores of three different grasping conditions.

The tactile sensor we used is TakkStrip [24] produced by Righthand Labs, which can only measure the pressure of 5 points simultaneously. To assess the quality of a grasp, we design a group of robot actions: 2 fast end-effector moving actions and 2 fast joint shaking actions. Before these actions, the tactile sensor measures the pressure first, which is the value in row 0. After each movement, the tactile sensor measures the value in time, and the values fill in the remaining four rows in a tactile image. Convolution calculation is applied to tactile images to extract features. And the three grasping conditions are:

- **Falling.** Fig. 3. (a) shows an example of the tactile image when the object falls. Obviously, falling means huge pressure changes at one contact point. We design a 1*2 convolution kernel (Fig. 3. (d)) to make convolutions with images. And the 1*2 convolution kernel is called as time kernel for the calculation process focuses on difference of the same point at different moments. The convolution result is a 4×5 matrix, and each row means the differences after one shake movement. And then we will find the row number where the maximum appears, which means the time of falling. The row number will be regarded as '$i$' and put into the 2nd line of eq.1.

- **Slippery.** Fig. 3. (b) shows an example of the tactile image when the object slips. The pressure values at adjacent points will change inversely when slip happens. For example, if the object slips from point 1 to point 2, the pressure value of point 1 will have a sharp reduce and point 2 will experience a sharp increase. The sum of the absolute value represents the magnitude of the slip. We design a 2*2 convolution kernel (fig. 3. (e)) to evaluate the slip of one shake movement, and it is called as space kernel for the function of calculating the difference about neighboring points. The convolution operation between space kernel and 5×5 tactile imprint actually calculates difference between the adjacent points of 4 shake movements, and the output is a 4×4 matrix. After that, we sum these 16 numbers together and put it into the third line of eq.1 (as '$x$'). This method can be also applied into the case where the contact points before and after slipping are not adjacent, such as slipping from point 1 to point 3.

- **Stable.** Fig. 3. (c) shows an example of the tactile image when the object is stable, and the score calculation shares the same process with 'slippery' states.

We summarize the grasp quality metric as eq.1:

$$\begin{cases} 0 & \text{If grasp fails,} \\ 0.5 \times \dfrac{i}{5} & \text{If falling occurs,} \\ 0.5 \times e^{-x/1000} + 0.5 & \text{If no falling is detected,} \end{cases} \quad (\text{eq.1})$$

where $i$ states the row number where the maximum value occurs after 2*1 kernel convolution operation, and $x$ is the calculation result after 2*2 kernel convolution with the tactile image.

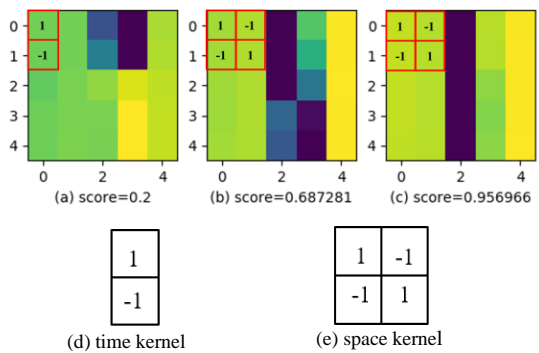

Fig. 3. Tactile images and convolution kernels

In this section, we define a metric for grasp quality with the continuous values from 0 to 1 rather than some simple discrete value existing in large prior papers. The 2 kernels and eq. 1 are mathematically explicable, which avoid the black box problem of hyper-parameters existing in neural networks. What's more, this method only requires a few contact points, rather than intensive spatio-temporal data acquisition, which can speed up the application of tactile sensing under the circumstance that the design and manufacture of tactile sensors are in the start-up stage.

### B. Prior tactile knowledge learning

We define the stability status as four categories: failure, falling, slippery and stable, with the score 0, (0, 0.5], (0.5, 0.85] and (0.85, 1]. Five objects (the known objects in Table I) are grasped 300 times to produce a dataset. Each image is rotated 180° as a data augmentation strategy, and a dataset including 3000 images and scores is formed, which is named as VPT dataset (see Fig. 1).

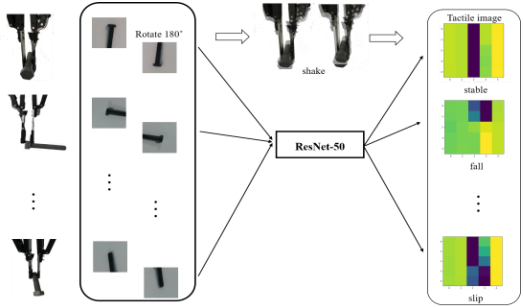

Fig. 4. Self-supervised tactile knowledge learning approach

Fig. 4. is the pipeline of the prior tactile knowledge learning approach. Before grasping, RGB images are captured, and during shaking, tactile data is recorded. ResNet-50 [25] is trained to build the connection between visual and tactile images, and the tactile data can be used to evaluate the grasp quality, and the labeling process is self-supervised based on the grasp quality metric mentioned in eq. 1. Therefore, we can obtain the grasp quality through a single RGB image. And we propose a strategy for data augmentation by rotating original images with 180°. The reason is that our gripper is 2-finger, image rotating 180° corresponds to the same grasp quality with the original image.

## IV. BAYESIAN GRASP

We propose Bayesian Grasp architecture in this section, and the whole process can be divided into two phases: grasp region estimation and stable grasp configuration generation. The first phase focuses on global perception, including target objects and background, which mainly relies on vision. The second phase aims to generate stable grasp configurations based on prior tactile knowledge learned from the training process, in other words, the target is realized with a score threshold of 0.85.

### A. Grasp region estimation phase

Undeniably, vision plays a significant role in robotic grasp detection. To manipulate multiple objects on the desk, the first step is to find a proper region that indicates a good grasp by visual sensing. A deep neural network is designed, whose architecture is shown in Fig. 5.

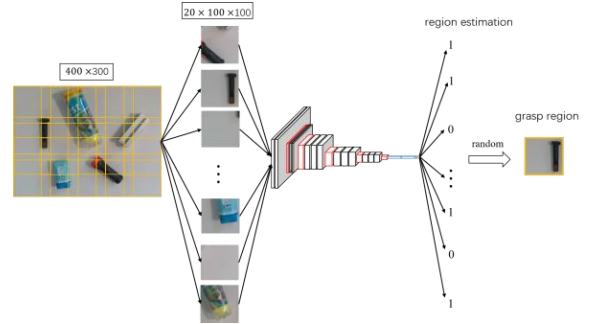

Fig. 5. The network architecture of grasp region estimation

**Network Design.** We design a bi-classification model $f(i)$ based on VGG-16 [26] network, consisting of 13 convolutional layers, 3 fully connected layers, 5 pool layers and a softmax layer with two outputs. The dataset is generated shown in the left picture of Fig.5. Given a global image of size 400×300, we select twenty image candidates by sliding window with stride of (80,78). The size of image candidate is 100×100 and its label is 1 or 0 based on whether the image contains an object.

During prediction phase, these 20 candidates are fed into network, and we select a candidate as potential graspable region from positive results randomly.

**Training.** To speed up the training process, we pre-train the network using weights from a model trained on ImageNet [27]. The VPT dataset is split into 2 categories, '0' (score=0) and '1' (score > 0), as the training dataset A. '1' means the candidate contains objects, which can be grasped with a proper grasp position and angle, and '0' means no graspable object. We train the model on this training dataset to minimize the loss $\Gamma(f, A) = \sum_{(i,s) \in A} L(f(i), s)$, where $L$ is the cross-entropy loss. We optimize the model with a batch size of 20 for 165 epochs and the learning rate is set as 0.01.

### B. Stable configuration generation phase

Based on the image candidate determined in grasp region estimation phase, a stable grasp configuration is generated through prior tactile knowledge. In this paper, the representation of grasp configuration is grasp position and grasp angle. Thus, the configuration is a 3-dimensional vector $(u,v,a)$, representing grasp pixel and grasp angle.

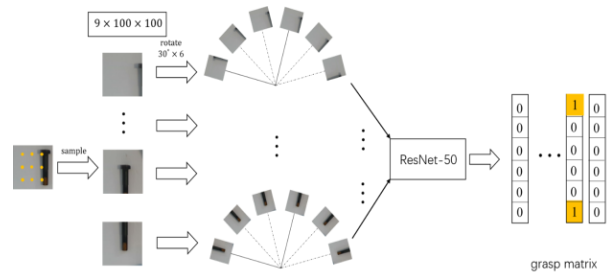

Fig. 6. The network architecture of stable grasp configuration

**Network Design.** We convert grasp angle regression as grasp classification. For an image candidate, nine image patches are extracted and rotated six times. Thus 54 image patches are generated totally representing different positions and angles. Each image represents a potential grasp configuration. We process each image patch using a convolutional network. Specially, a bi-classification model $g(i)$, as Fig. 6 shows, is designed based on 50-layer deep residual network. All the image patches are fed into ResNet-50, and 54 outputs are generated, corresponding to 54 potential grasp configurations. Positive outputs mean stable grasp points and angles.

**Training.** The weights pre-trained on ImageNet are used to speed up the training process. The VPT dataset is split into 2 categories, '0' (score $\leq 0.85$) and '1' (score $>0.85$), as the training dataset B. We train the model on this training dataset to minimize the loss $\Gamma(g, B) = \sum_{(i,s) \in A} L(g(i), s)$, where $L$ is the cross-entropy loss. We optimize the model with a batch size of 25 for 165 epochs and the learning rate is set as 0.01.

## V. EXPERIMENTS

To validate our method, a series of experiments has been designed to: i) verify the accuracy of our hand-design grasp quality metric, ii) evaluate the classification accuracy of prior tactile knowledge approach on known objects and generalization capability on novel test objects. Moreover, we test the Bayesian Grasp policy on real robot system to grasp known and novel objects.

### A. Grasp quality metric

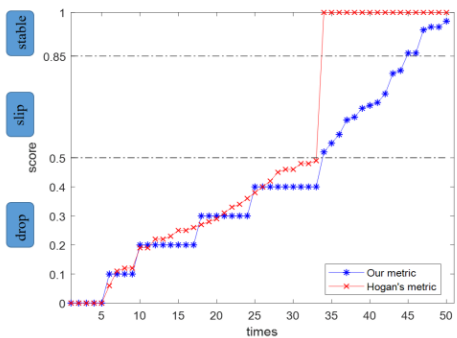

Fig. 7. Performance of our grasp quality metric

To test the effectiveness of the grasp quality metric, we did some benchmark experiments with the grasp evaluation proposed by Hogan et al. [6].

In the benchmark experiments, the robot didn't stop shaking after the predefined 4 movements but continued to shake until the object drops or shaking for 300 seconds. The total shaking time was recorded and divided by 300, which could be called as Hogan's metric. And then we grasped 50 times of 5 known objects. As is described in Fig. 7, our metric has great consistency with Hogan's metric, while our metric is more detailed at the part of slip detection.

### B. Prior tactile learning approach

**Performance on known objects.** We obtain a training accuracy of 92% and a testing accuracy of 86%. This result shows that given a grasp about a known object, we can get a stable grasp configuration reliably.

**Performance on unknown objects.** The VPT dataset is split into 2 categories: '0' (score $\leq 0.85$) and '1' (score $> 0.85$). And the classification accuracy for each object is 66% (screw), 82% (yogurt), 85% (tennis container), 77% (flashlight) and 83% (metal bar). The results show that this method has the ability to generalize to other new objects.

### C. Hardware setup

As Fig. 8 shows, in this experiment, we build a hardware platform consisting of a 6-DoF UR5 arm from Universal Robots, the 2-finger adaptive robot gripper from Robotiq, Realsense D435 RGB-D camera and a TakkStrip produced by Takktile. Among them, the camera is mounted on the top of the desk to provide visual data. The TakkStrip is a tactile sensor which can measure pressures of 5 contact points on the surface simultaneously. It consists of 5 sensors end-to-end spaced at 8mm spacing.

The network training is implemented in PyTorch on a server, consisting of a 4.2 GHz Intel Core i7-7700HQ CPU (four physical cores), 16GB of system memory, and a GTX 1080Ti GPU.

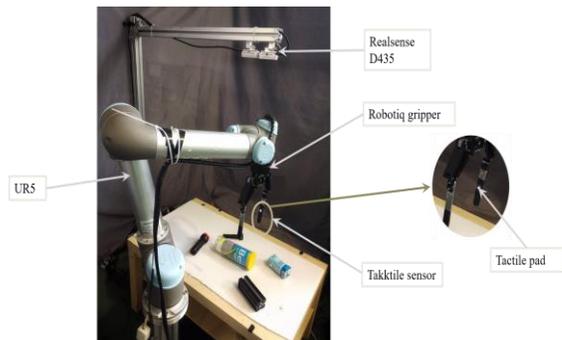

Fig.8. Setup of the robot system

### D. Comparison with Vision-based Grasp in robot system

To evaluate the performance of our proposed framework, we retrain a Vision-based grasp model sharing the same structure with $g(i)$ model designed in Section IV.B, however, with new labels acquired only by visual information. Clearly, vision can easily detect drop but hardly detect slip, so the grasps in VPT dataset are labelled as '1' (score>0.5) and '0' (score$\leq 0.5$) according to the score. We optimize the model with a batch size of 25 for 165 epochs and the learning rate is set as 0.01, which is as same as Section IV.B.

We grasp 5 known and 5 novel objects in real-time with a real robot hand based on Bayesian Grasp and Vision-based Grasp strategy. As figure 8 shows, 5 objects are randomly placed in a size of 400 by 300 in the center of the table without any shading.

TABLE I.     PRIOR TACTILE KNOWLEDGE IMPROVES GRASPING ACCURACY

| Objects | Known Objects | | | | | Novel Objects | | | | |
|---|---|---|---|---|---|---|---|---|---|---|
| | Screw | Yogurt | Tennis container | Flashlight | Metal bar | Lay's | Realsense box | Shaving foam | Guangming | Unity knife |
| Weight | 292.52g | 211.72g | 221.10g | 76.52g | 222.62g | 128.48g | 130.90g | 134.36g | 304.29g | 71.25g |
| Size (axial direction) | 112mm | 100mm | 210mm | 131mm | 130mm | 210mm | 145mm | 130mm | 142mm | 130mm |
| Images | 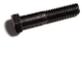 | 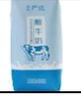 | 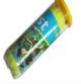 | 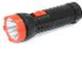 | 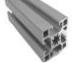 | 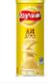 | 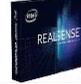 | 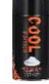 | 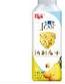 | 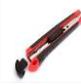 |
| *Vision-based Grasp* | 34% | 64% | 52% | 70% | 42% | 62% | 54% | 66% | 32% | 70% |
| *Bayesian Grasp* | 90% | 92% | 76% | 96% | 88% | 70% | 86% | 88% | 76% | 94% |
| *Relative improvement* | 165% | 44% | 46% | 37% | 110% | 13% | 59% | 33% | 138% | 34% |

The rates of successful grasps (score>0.85) are listed in Table I. It's easy to draw the conclusion that the introduction of prior tactile knowledge increases the overall accuracy by an average of 36% on known objects and 26% on novel objects.

## VI. EXPERIMENTS

This paper proposes a novel framework to improve grasp stability by using prior tactile knowledge. Results on a test set of 5 known objects and 5 novel objects show that our method yields average relative improvement of the accuracy of 50% over traditional vision-based strategies in the same grasping time. What's more, the grasping process displays anthropomorphic intelligence completely and hints the potential future of stable grasp based on tactile sensing.

**Discussion on the experiment results.** The grasp quality metric proposed in this paper shows good performance on quality evaluation, especially for slip detection. And the prior tactile learning method can be generalized within a certain range. Actually, the method breaks down the gap between modalities and transfers the shared features based on cross-modal learning. However, the difference of accuracy between different objects shows that the diversity of shape and weight bring great challenges.

The Bayesian Grasp shows great improvement to heavy and out-of-shape objects in real robot experiments over the accuracy of Vision-based, especially for heavy objects. Personally, the key point is that external force is introduced, which put forward higher demands on precise stable grasp point and angle. And Bayesian Grasp policy is effective to generate stable grasp configuration under the help of learned prior tactile knowledge.

**Future work.** The tactile information is dynamic and continuous. Analyzing and Learning the changes among continuous tactile information would better imitate humans' manipulation and improve the stability during the manipulation process. Besides, the tactile sensors we used can only sample limited contact points in a straight line. Designing a flexible tactile sensor with high resolution might be better to reflect the contact situation in detail. What's more, a set of tactile standard on acquisition, transmission, and processing might be built up to speed up the current initial research, and the study of computer vision might be referred.

## ACKNOWLEDGMENT

This research is supported by Special Program for Innovation Method of the Ministry of Science and Technology, China (2018IM020100), National Key Scientific Instruments and Equipment Development Program of China (2016YFF0101602, 2016YFC0104104), National Natural Science Foundation of China (51775332, 51675329), and the Cross Fund for medical and Engineering of Shanghai Jiao Tong University (YG2017QN61).